\documentclass[conference, letterpaper, 10 pt]{ieeeconf}  

\IEEEoverridecommandlockouts                              
\usepackage[utf8]{inputenc}
\usepackage[T1]{fontenc}
\usepackage{amsmath,amssymb,mathrsfs}
\usepackage{graphicx}
\usepackage{algorithm}
\usepackage[noend]{algpseudocode}
\usepackage{mdwtab}
\usepackage{eqparbox}
\usepackage{url}
\usepackage[colorlinks=true, allcolors=blue]{hyperref}
\usepackage[usenames,dvipsnames,svgnames]{xcolor}
\usepackage{mathtools}
\usepackage{array}
\usepackage{booktabs}
\usepackage{siunitx}
\usepackage{textgreek}
\usepackage{comment}
\usepackage{leftidx}
\usepackage{subfigure}
\usepackage{amsfonts}
\usepackage{tikz}
\usepackage{graphicx}

\newcommand{\method}[0]{VIHE }

\title{\LARGE \bf
VIHE: Virtual In-Hand Eye Transformer for 3D Robotic Manipulation}

\author{Weiyao Wang$^{1,\dagger}$, Yutian Lei$^{2}$, Shiyu Jin$^{2}$, Gregory D. Hager$^{1}$ and Liangjun Zhang$^{2}$
\thanks{$^{1}$ W. Wang and G. Hager are with the Johns Hopkins University, Department of Computer Science, Baltimore, USA.
        {\tt\small wwang121 @ cs.jhu.edu \text{and} hager @ cs.jhu.edu}
        }
\thanks{$^{2}$ Y. Lei, S. Jin, L. Zhang are with Baidu Research, Robotics and Autonomous Driving Lab (RAL), Sunnyvale, USA.
        {\tt\small \{yutianhe,shiyujin,liangjunzhang\} @ baidu.com}
        }
\thanks{$^{\dagger}$ Work done while the author was an intern at Baidu Research, Robotics and Autonomous Driving Lab (RAL), Sunnyvale, USA.
        }
}

\begin{document}
\maketitle


\begin{abstract}
In this work, we introduce the Virtual In-Hand Eye Transformer (VIHE), a novel method designed to enhance 3D manipulation capabilities through action-aware view rendering. VIHE autoregressively refines actions in multiple stages by conditioning on rendered views posed from action predictions in the earlier stages. These virtual in-hand views provide a strong inductive bias for effectively recognizing the correct pose for the hand, especially for challenging high-precision tasks such as peg insertion. On 18 manipulation tasks in RLBench simulated environments, VIHE achieves a new state-of-the-art, with a 12\% absolute improvement, increasing from 65\% to 77\% over the existing state-of-the-art model using 100 demonstrations per task. In real-world scenarios, VIHE can learn manipulation tasks with just a handful of demonstrations, highlighting its practical utility. Videos and code implementation can be found at our project site: \url{https://vihe-3d.github.io}.
\end{abstract}

\section{Introduction}
Achieving mastery in manipulating objects in 3D environments is a foundational goal in the quest for intelligent real-world robotic systems. While machine learning and computer vision have propelled significant advancements in robotic manipulation~\cite{argall2009survey, kroemer2021review,ahn2022can,huang2022language,driess2023palm}, the challenge of crafting an effective observation space for effective learning in 3D persists. Current methodologies in the realm of robotic manipulation employ various types of 3D representations. Recently in the space of end-to-end vision-based imitation learning in 3D, PerAct~\cite{shridhar2023perceiver} utilizes voxel-based representations, which, although powerful, suffer from computational inefficiencies due to the cubic scaling of voxels. RVT~\cite{goyal2023rvt} employs multi-view orthographic images but faces difficulties in tasks that demand high-precision 3D reasoning. Act3D~\cite{gervet2023act3d} leverages point clouds for 3D representation but also computationally suffers from large number of sampled points and neglects the potential advantages of spatial biases in manipulation tasks.

\begin{figure}[h!]
  \centering 
  \includegraphics[width=3.4in]{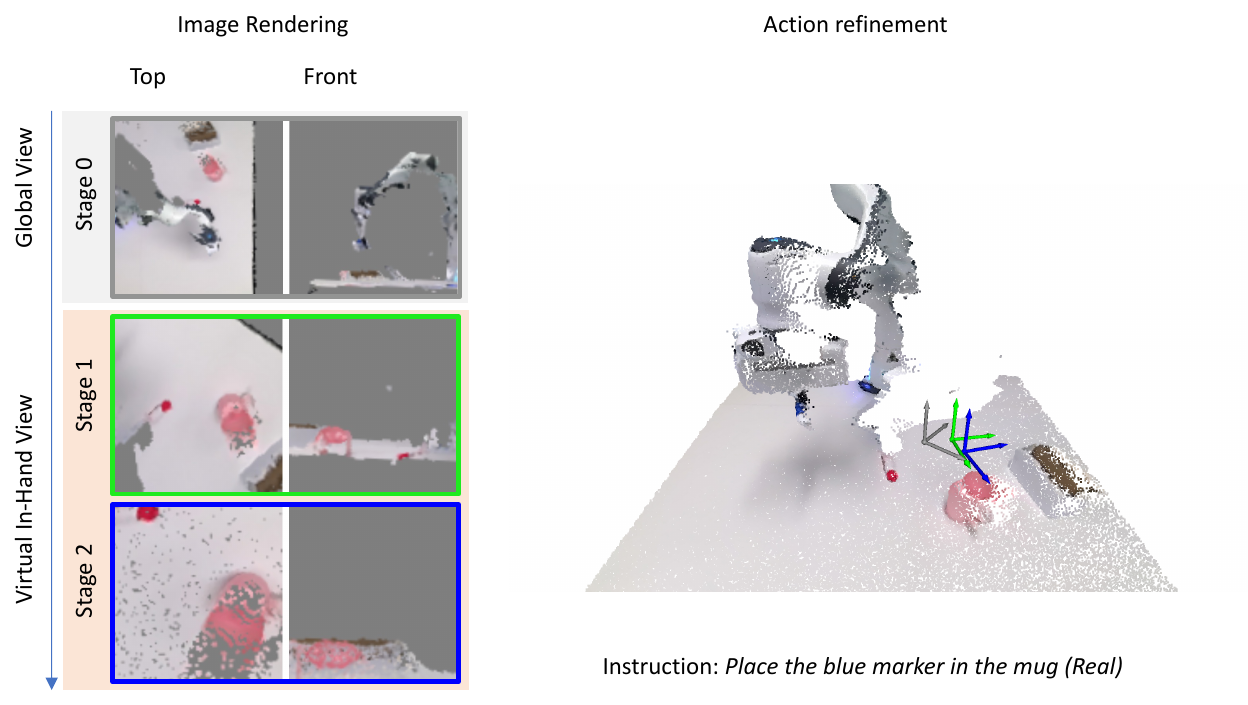}
  \caption{Visual example of \method in real-world. Our method iteratively refines its 3D action prediction (right) by rendering 2D in-hand views based on the previous stage predictions (left). Color coding of gray, green, and blue represent three action prediction stages respectively.}
  \label{fig:teaser}
\end{figure}

\begin{figure}[t!]
  \centering 
  \includegraphics[width=3.4in]{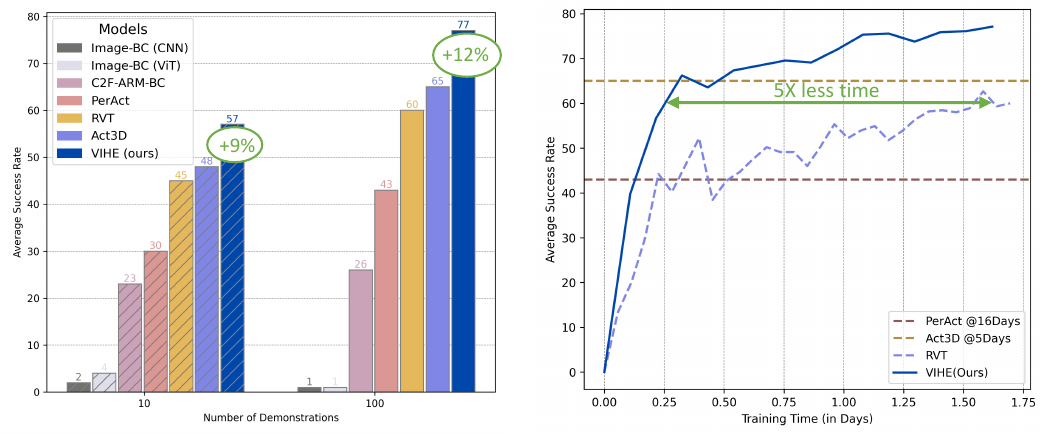}
  \caption{\method scales and performs better than RVT, PerAct, and Act3D among other baselines. Attribute to the inductive bias from in-hand views, \method also require 5X less time to achieve on-par performance to the previous SOTA method. }\label{fig:ret_summary}
\end{figure}

\begin{figure*}[h!]
  \centering 
  \includegraphics[width=7in]{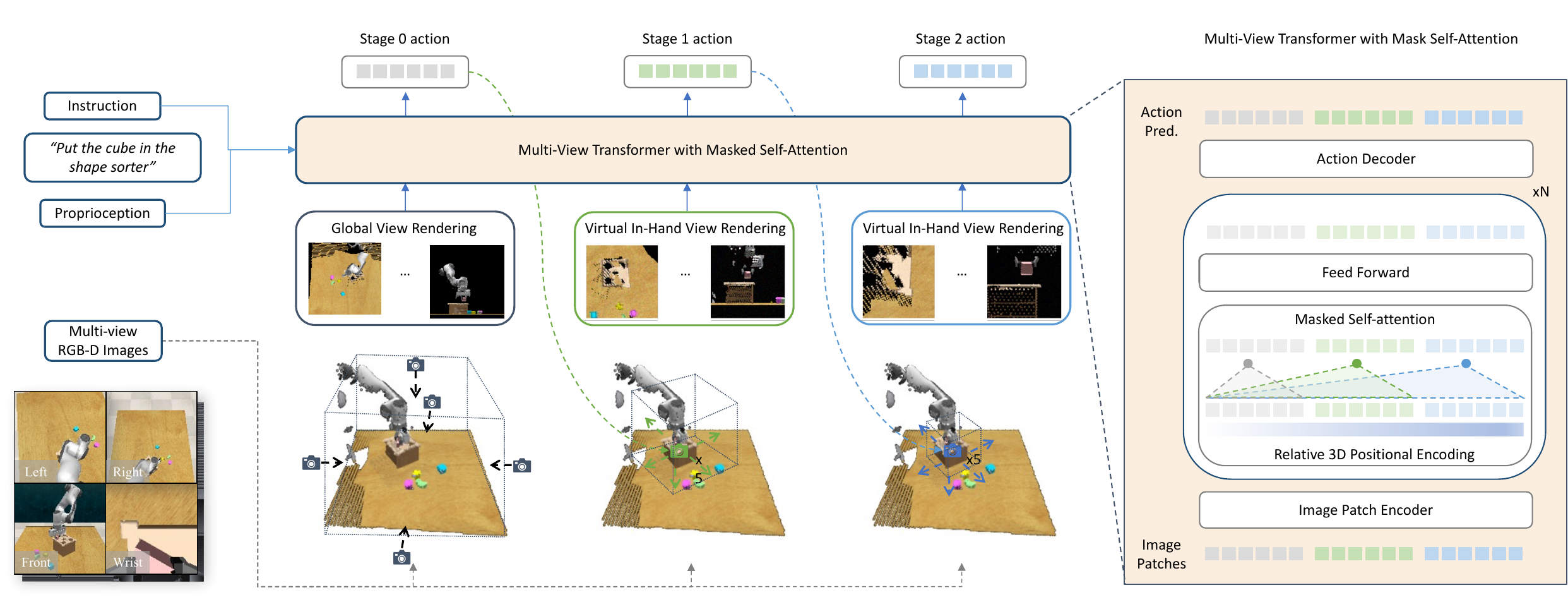}
  \caption{\textbf{\method Overview.} Starting with RGB-D images from multi-view cameras, we first construct a point cloud of the scene. Global views are first rendered using fixed cameras positioned around the workspace. From these global views, the network outputs initial action predictions $a_{pose}^0, a_{open}^0, a_{col}^0$. Then at each refinement stage $i$, we autoregressively generate virtual in-hand views from cameras attached to the previously predicted gripper pose $a_{pose}^{i-1}$. Based on the rendered views, we then refine the action predictions. The network architecture employs masked self-attention to have tokens from later stages attend to tokens from previous stages. Language instruction tokens are merged into stage 0 image tokens when input into transfer, which is omitted in the figure for conciseness. More information can be found in Sec.~\ref{sec:method}.}\label{fig:architecture} 
\end{figure*}

We observe that existing approaches often treat the 3D workspace uniformly, neglecting the naturally occurring inductive bias that the space near the end-effector holds significant importance for manipulation tasks. Previous research has underscored the value of an in-hand perspective: for instance, \cite{jangir2022look} demonstrates that an in-hand view reveals more task-relevant details, which is particularly advantageous for high-precision tasks. Similarly, \cite{hsu2022vision} shows that incorporating an in-hand view can mitigate distractions irrelevant to gripper actions, thereby improving generalization. However, these studies are primarily limited to 2D image-based manipulation~\cite{jangir2022look,hsu2022vision}, utilizing a physical camera attached to the robot's end-effector. Consequently, we aim to extend the inductive bias of the end-effector location to better structure observations for 3D manipulation tasks.

We introduce the Virtual In-Hand Eye Transformer (\method) which utilizes an iterative process to refine action predictions, leveraging rendered in-hand views at each stage. In each refinement stage, we render local images based on the predicted action in the previous stage. Conditioned on these in-hand perspectives, the model subsequently predicts relative SE(3) transformations to refine the end-effector pose from the previous stage. The action pose is then updated by applying this refinement transformation. This iterative procedure can be executed across multiple stages to enhance accuracy. A visual example of this process can be found in Figure~\ref{fig:teaser}. Intuitively, we are asking the model solve a local action refinement problem conditioned on local image context. Such local image context, other than providing higher effective resolution with same image size, is also shown by~\cite{hsu2022vision} to be beneficial for generalization as less distracting information is provided to the model. Throughout these stages, we employ masked self-attention mechanisms, allowing tokens from later stages to attend to those from earlier stages and blocking attention in the opposite direction during training. This design effectively integrates information from rendered images across all stages, leading to more accurate final predictions. 

Our empirical evaluations affirm the efficacy of incorporating virtual in-hand perspectives into 3D robotic manipulation tasks. We demonstrate that our approach significantly enhances performance across a variety of tasks and settings, thereby providing an effective solution for real-world applications. As shown in Figure~\ref{fig:ret_summary}, our method delivers an 18\% improvement in final performance and requires only one-fifth of the training time to achieve performance metrics comparable to existing state-of-the-art methods. In high-precision tasks such as peg insertion, our approach more than triples the success rate when compared to current state-of-the-art methods. Conducted on a Franka robot, we also validate the superiority of our method over baseline approaches through real-world experiments. Lastly, through extensive ablation studies, we validate our design choices, such as rendering from in-hand perspective, zoom-in local image rendering and action prediction by stage-wise refinement, all contribute to final performance improvements.

Our contributions can be summarized below.
\begin{enumerate}
    \item We introduce a novel representation technique that utilizes virtual in-hand views, and employ a transformer-based network architecture to iteratively refine action predictions based on this representation.
    \item We investigate various design choices to efficiently utilize this representation in robotic manipulation tasks. 
    \item Through empirical evaluation in both simulated and real-world settings, we demonstrate significant improvements in training speed, sample efficiency, and overall performance.
\end{enumerate}

\section{Related Work}

\subsection{Vision-Based Imitation Learning for Robotic Manipulation}

The landscape of imitation learning~\cite{zheng2022imitation} in robotic manipulation has undergone a significant transformation with the advent of deep learning. Traditional approaches often depended on low-dimensional state observations and engineered features~\cite{kemp2007challenges, bandera2012survey, murray2017mathematical}. With the advancement of deep learning~\cite{lecun2015deep,goodfellow2016deep}, studies have explored using convolutional neural networks~\cite{gu2018recent} to have raw 2D RGB images as inputs and predict actions directly~\cite{codevilla2018end,wang2019learning,pari2021surprising,young2021visual}. More recently, works such as RT-1~\cite{RT1}, RT-2~\cite{RT2}, GATO~\cite{GATO} and Octo~\cite{octo_2023} have employed Transformer architectures~\cite{vaswani2017attention,lin2022survey} to predict robot actions directly from high-dimensional multi-modal inputs like images and natural language instructions. However, these methods typically require a large amount of demonstrations for effective learning~\cite{RT1, RT2, GATO, octo_2023}.

The field has also experienced a growing interest in the use of RGB-D images as input, offering richer visual information in the 3D space where manipulation occurs. Methods like CLIPort~\cite{CLIPort} have shown promise in simple pick-and-place tasks but remain confined to tabletop, top-down scenarios. To fully exploit the 3D information available in observations, research efforts such as C2F-ARM~\cite{C2FARM} and PerAct~\cite{PerAct} have utilized 3D voxel grids to represent the robot's workspace. While these methods offer spatially precise 3D pose prediction, they come with the drawback of high computational overhead.

Among the most recent advancements~\cite{goyal2023rvt,gervet2023act3d,parashar2023spatial,ze2023gnfactor,ke20243d}, the state-of-the-art method Act3D~\cite{gervet2023act3d} extracts point clouds from RGB-D images and characterizes them using pre-trained CLIP~\cite{radford2021learning} features. Actions are subsequently identified from additionally sampled action proposal points through cross-attention with the previous feature points. Due to the large number of points that need to be sampled for both features and action proposals, Act3D remains computationally demanding, requiring five days of training with only two cross-attention layers. Another recent work, RVT~\cite{goyal2023rvt} is the most closely related to our own. Like Act3D, RVT also extracts point clouds from RGB-D images but opts to render multi-view orthographic images~\cite{szeliski2022computer} using these points. The orthographic images provide a useful inductive bias for action location prediction, enabling RVT to achieve performance similar to Act3D but with a significantly reduced training time of just two days. Our method also employs orthographic images as observational input but distinguishes itself by utilizing in-hand views to iteratively refine action pose predictions, setting it apart from RVT and other prior methods.

\subsection{In-Hand View for Robotic Manipulation}

Previous research has also highlighted the advantages of an in-hand perspective for robotic manipulation that requires high precision~\cite{chi2024universal, jangir2022look,hsu2022vision,puang2020kovis,song2020grasping,valassakis2021coarse}. For instance, \cite{jangir2022look} demonstrates that an in-hand view reveals more task-relevant details, which is particularly beneficial for high-precision tasks such as peg insertion. Besides precision, \cite{hsu2022vision} also shows that incorporating an in-hand view can mitigate distractions that are irrelevant to gripper actions, thereby improving generalization. Furthermore, ~\cite{chi2024universal} builds a data collection system that can transfer human demonstrations to robot policies, leveraging in-hand camera views. However, these studies typically obtain the in-hand perspective through a camera attached to the robot's end-effector, which differs from our approach of rendering in-hand views from point clouds. We further argue that directly using camera-captured in-hand perspectives—when available—would not be as effective in our keypoint-based imitation learning setting, to be elaborated in Sec.~\ref{sec:method}. This is because the next predicted keypoint could be located at a considerable distance from the current hand position (such as in Figure~\ref{fig:teaser}), rendering the existing in-hand view less useful.

\section{Methods}\label{sec:method}

\subsection{Overview}
We investigate the impact of in-hand camera views in the setting of language-conditioned imitation learning with a 3D action and observation space. We assume access to a dataset comprising \( n \) demonstration trajectories. Each demonstration consists of a language instruction \( l \), a sequence of $m$ observations $\{^{0}o,\hdots,^{m-1}o\}$  and a sequence of $m$ continuous actions $\{^{0}a,\hdots,^{m-1}a\}$. In this work, we treat each time step independently, so for simplicity of notation in subsequent developments, we consider a single time step and we use $o$ and $a$ to represent the observation and action at that time. Each observation \( o \) includes RGB-D images captured from one or more camera perspectives. An action \( a \) consists of a 6-DoF end-effector pose \( a_{\text{pose}} \) in SE(3), a binary open or closed state \( a_{\text{open}} \), and a binary state indicating whether to employ a motion planner to avoid collisions \( a_{\text{col}} \) while moving to the target pose:
\begin{align}
    a &= \{ a_{\text{pose}} \in \text{SE}(3),  a_{\text{open}}\in \{0,1\},  a_{\text{col}}\in\{0,1\}\}. 
\end{align}

\begin{figure*}[h]\label{fig:views}
  \centering 
  \includegraphics[width=6.25in]{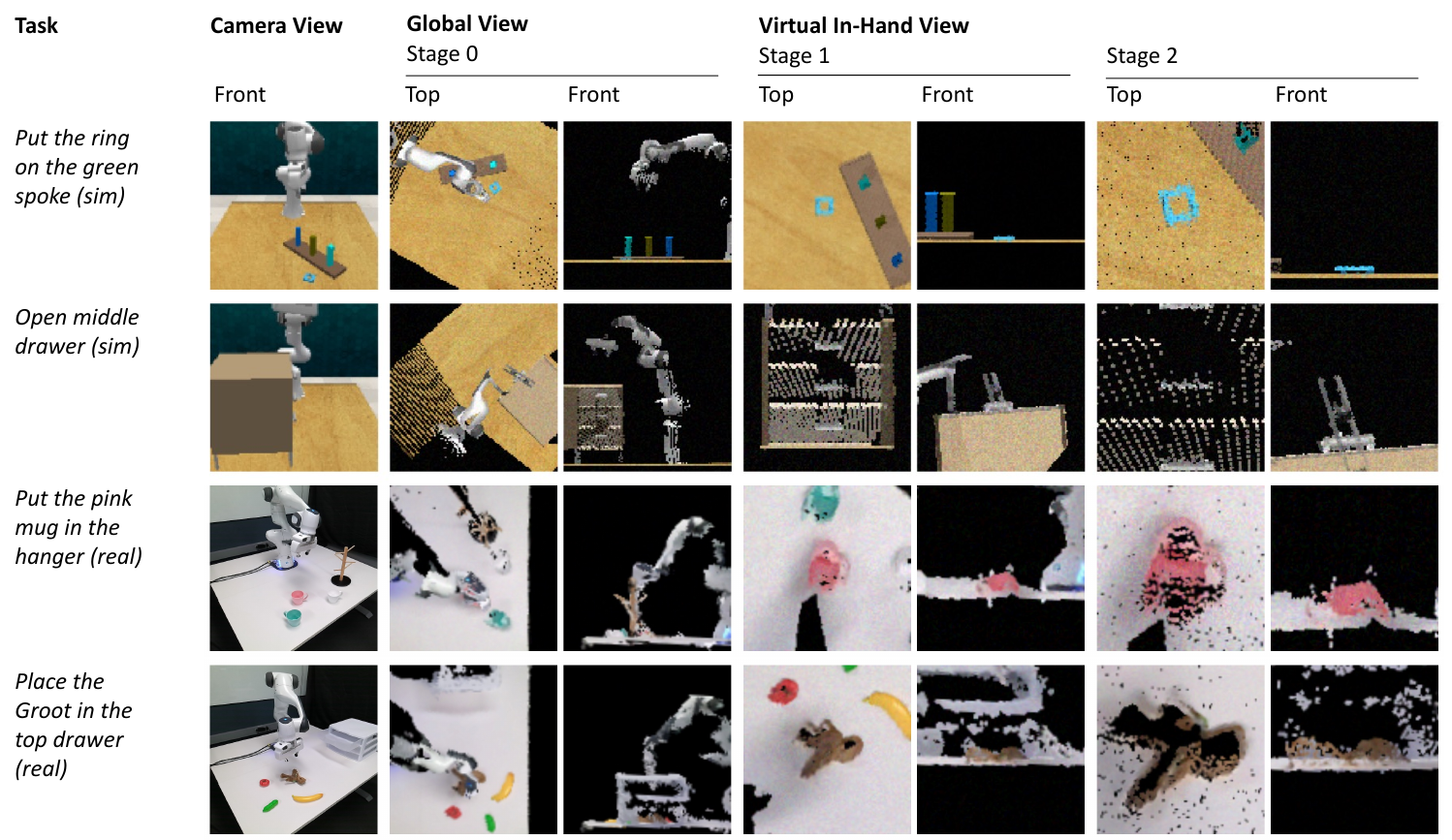}
  \caption{Visualization of sample images from front camera in RGB view, global view and virtual in-hand view (both rendered as orthographic images). For global orthographic view and virtual in-hand view, two (top and front) out of five views (top, front, back, left and right) are visualized. Virtual in-hand view reveals information that is occluded from global view and in greater details, leading to better manipulation performance.}
\end{figure*}
The architecture of the Virtual In-Hand Eye Transformer (VIHE) is depicted in Figure~\ref{fig:architecture}. It employs a Transformer-based policy that, at given timestep \( t \), predicts an end-effector pose in SE(3) based on one or more RGB-D images, a language instruction, and proprioceptive state of the robot's current end-effector state. The core concept is to predict and refine the end-effector pose using images rendered according to the predicted pose produced at each stage in an autoregressive process. In line with prior work~\cite{goyal2023rvt,gervet2023act3d,PerAct}, 
we identify a set of ``keyposes'' that capture critical end-effector poses within a demonstration and apply VIHE at only those steps. A pose qualifies as a keypose if: (1) the end-effector changes state (e.g., something is grasped or released), (2) velocities approach zero (commonly observed when entering pre-grasp poses or transitioning to a new phase of a task). The prediction problem is thus reduced to identifying the next action at each keypose based on the current observation. During inference, \method iteratively predicts the next best keypose and navigates to it using a motion planner, as in previous works~\cite{goyal2023rvt,gervet2023act3d,PerAct}.

\subsection{Iterative View Rendering and Action Refinement} \label{sec:method_refine}

In the following, we describe the multi-stage view rendering and action refinement procedure that iteratively predicts and refines action using the multi-view transformer network, $F$. A visualization of the process can be found in Figure~\ref{fig:architecture}.

\subsubsection{Initial Global Stage}
The initial global stage (Stage 0)  utilizes a predefined set of fixed virtual camera poses \( p^0 \) positioned around a cubic workspace from five directions (top, front, back, left, and right) as in RVT~\cite{goyal2023rvt}. These poses are accompanied by intrinsic parameters \( f^0 \) chosen so that each rendered image that covers the size of the entire workspace. Using a rasterization-based renderer \(R(o, p^0, f^0)\), we generate images \( x^0 \) from these camera poses and intrinsics. The multi-view transformer network \( F \) then predicts the action \( a^0_{\text{pose}}, a^0_{\text{open}}, a^0_{\text{col}} \) based on these images \( x^0 \) and a given language instruction \( l \). This can be expressed as:
\begin{align}
    x^0 &= R(o, p^0, f^0), \\
    a^0_{\text{pose}}, a^0_{\text{open}}, a^0_{\text{col}}  &= F(x^0, l). \label{eq:init}
\end{align}

\subsubsection{Iterative Refinement}
In subsequent stages, we generate virtual in-hand view images to provide input for iterative action refinement. These images are rendered from five cameras with the same relative geometry, but now placed relative to the predicted end-effector pose, effectively serving as "virtual in-hand eyes." To focus on finer details, the intrinsic parameters \( f^i \) are adjusted to cover \( \frac{1}{2^i} \) of the workspace at each stage \( i \). 
From $[x^0,\hdots,x^i]$ (all rendered images up to stage i) along with language instruction $l$, we use the network $F$ to predict a transformation refinement $h_{pose}^i$ and updated $a^i_{\text{open}}, a^i_{\text{col}}.$ The refinement $h_{pose}^i$ is then applied to the previously predicted action pose $a^{i-1}_{pose}$ to obtain an updated action pose in stage $a^{i}_{pose}$. We chose to predict a relative transformation instead of the pose directly to encourage the network to weight information on in-hand views over fixed views in the refinement stages. An ablation study supporting this choice is provided in Sec.~\ref{sec:sim_exp}. Let \( C \) be a function that maps action poses to camera poses. The iterative refinement can be formalized as:
\begin{align}
    p^i &= C(a_{\text{pose}}^{i-1}), \\
    x^i &= R(o, p^i, f^i), \\
    h_{pose}^i, a^i_{\text{open}}, a^i_{\text{col}}  &= F([x^0,\hdots,x^i], l), \label{eq:refine} \\ 
    a^i_{\text{pose}} &= h_{pose}^i \circ a^{i-1}_{\text{pose}}.
\end{align}

Note that Eq.~\ref{eq:init} can be regarded as a special case of Eq.~\ref{eq:refine} where we refine from base coordinate frame as an intial starting pose. 

\subsection{VIHE Architecture}

\subsubsection{Network Architecture}

The rendered images $[x^0,\hdots,x^2]$, the task language instruction $l$, and the proprioception data are processed by a transformer network $F$. In order to ground the instruction in the scene,  we use a pre-trained CLIP~\cite{CLIP} model to extract a sequence of $K_{lang}$ tokens per instruction. For both fixed and in-hand views, we break each rendered image into $K_{img}$ patches. A one-layer convolutional layer is used to extract in total \(5\times K_{img}\) tokens for each stage that renders 5 views. For the proprioception data, similar to PerAct and RVT, we pass it through an MLP and concatenate it to the image tokens. We repeat refinement two times - we have 1 initial global stage and 2 refinement stages.

To efficiently train action prediction for all stages, \method has $L$ masked self-attention layers~\cite{vaswani2017attention} following the patch encoding layer. Given the set of all \(3\times5\times K_{img} + K_{lang}\) tokens across all 3 stages, we construct a mask \( M \) so that each token can attend to other tokens within the same or earlier stages. Language tokens are treated as stage 0 tokens to allow all stage tokens to access instruction information. 
The masked self-attention is then formulated as:
\begin{equation}
    F_{\text{attention}} = \text{Softmax}\left(\frac{QK^T + M}{\sqrt{d_k}}\right) V
\end{equation}
where $Q,K,V$ denotes query, key and value vectors in dimension $d_k$ in each attention layer.

To efficiently fuse information from multi-views captured from different camera poses in 3D, we apply the Rotary Position Embedding (RoPE)~\cite{RotaryEmbeddings} to the query and key when calculating the attention scores. For each image patch token, we compute its position projected to the camera image plane and use that 3D location to compute RoPE embedding. This is similar to Act3D~\cite{gervet2023act3d} where they apply RoPE to cross-attention over points. For all language tokens, we directly learn a RoPE embedding function since no location is relevant to the language. We additionally add learned positional embedding to language tokens and learned view-specific embedding to image tokens.

\begin{table*}[!t]
    \caption{Multi-Task Performance on Simulated Benchmark.}
    \label{tab:rlbench_main_ret}
    \centering
    \scriptsize
    \begin{tabular}{ccccccccccccccccccccccccccccccccccccccc}
        \toprule
         & \multicolumn{2}{c}{Avg.} & \multicolumn{2}{c}{Training} & \multicolumn{2}{c}{Close} & \multicolumn{2}{c}{Drag} & \multicolumn{2}{c}{Insert} & \multicolumn{2}{c}{Meat off} & \multicolumn{2}{c}{Open} & \multicolumn{2}{c}{Place} & \multicolumn{2}{c}{Place} & \multicolumn{2}{c}{Push} \\
        Models & \multicolumn{2}{c}{Success} & \multicolumn{2}{c}{Time (in Days)} & \multicolumn{2}{c}{Jar} & \multicolumn{2}{c}{Stick} & \multicolumn{2}{c}{Peg} & \multicolumn{2}{c}{Grill} & \multicolumn{2}{c}{Drawer} & \multicolumn{2}{c}{Cups} & \multicolumn{2}{c}{Wine} & \multicolumn{2}{c}{Buttons} \\
        & 10 & 100 & & &10 & 100 &10 & 100 & 10 & 100& 10 & 100& 10 & 100&10 & 100&10 & 100&10 & 100&\\
        \midrule
        Image-BC (CNN) & 2 & 1 & \multicolumn{2}{c}{-} & 0 &0 & 0 &0 & 0 &0 & 0 &0 & 4 &4 & 0 &0 & 0 &0 & 4 &0 \\
        Image-BC (ViT) & 4 & 1 & \multicolumn{2}{c}{-} & 0 &0 & 0 &0 & 0 &0 & 0 &0 & 16 &0 & 0 &0 & 4 &0 & 16 &0 \\
        C2F-ARM-BC & 23 & 26 & \multicolumn{2}{c}{-} & 28 &24 & 72 &24 & 0 &4 & 40 &20 & 28 &20 & 0 &0 & 36 &8 & 88 &72 \\
        PerAct & 30 &43 & \multicolumn{2}{c}{16} & 32 &60 & 36&68 & 4& 0 & 68&84 & 68&80 & 0&0 & 20&12 & 56&48 \\
        RVT & 45 & 60 & \multicolumn{2}{c}{1.7} & 52 & 44 & \textbf{100} & \textbf{100} & 8 & 12 & \textbf{72} & 92 & 84 & 76 & 0 & 4 & 60 & 56 & 72 &\textbf{100} \\
        Act3D & 48 & 65 & \multicolumn{2}{c}{5} & \textbf{90} & \textbf{92} & 52 &92 & 7 &27 & 58 &94 & \textbf{92} & \textbf{93} & 1 &3 & 32 &80 & \textbf{98} &99 \\
        VIHE (ours) & \textbf{57} &\textbf{77} & \multicolumn{2}{c}{\textbf{1.6}} & 36 &48 &  \textbf{100} &\textbf{100} & \textbf{24} &\textbf{84} & \textbf{72} &\textbf{100} & 72 &76 & \textbf{4} & \textbf{12} & \textbf{60} & \textbf{88} & 88 &\textbf{100} \\
        \midrule
         & \multicolumn{2}{c}{Put in} & \multicolumn{2}{c}{Put in} & \multicolumn{2}{c}{Put in} & \multicolumn{2}{c}{Screw} & \multicolumn{2}{c}{Slide} & \multicolumn{2}{c}{Sort} & \multicolumn{2}{c}{Stack} & \multicolumn{2}{c}{Stack} & \multicolumn{2}{c}{Sweep to} & \multicolumn{2}{c}{Turn} \\
        Models & \multicolumn{2}{c}{Cupboard} & \multicolumn{2}{c}{Drawer} & \multicolumn{2}{c}{Safe} & \multicolumn{2}{c}{Bulb} & \multicolumn{2}{c}{Block} & \multicolumn{2}{c}{Shape} & \multicolumn{2}{c}{Blocks} & \multicolumn{2}{c}{Cups} & \multicolumn{2}{c}{Dustpan} & \multicolumn{2}{c}{Tap} \\
        & 10 & 100 & 10 & 100& 10 & 100 &10 & 100 & 10 & 100& 10 & 100& 10 & 100&10 & 100&10 & 100&10 & 100&\\
        \midrule
        Image-BC (CNN) &0 & 0 &0& 8 &0& 4 &0& 0 &4& 0 &0& 0 &0& 0 &0& 0 &0& 0 &20& 8 \\
        Image-BC (ViT) &4& 0 &0& 0 &0& 0 &0& 0 &8& 0 &0& 0 &0& 0 &0& 0 &8& 0 &24& 16 \\
        C2F-ARM-BC &4& 0 &12& 4 &0& 12 &12& 8 &12& 16 &8& 8 &4& 0 &0& 0 &4& 0 &60& 68 \\
        PerAct &0& 16 &16& 68 &16& 44 &28& 24 &32& 72 &16& 20 &12& 36 &0& 0 &72& 56 &72& 80 \\
        RVT & 16 & 52 & 60 & 92 &44& 56 &16& 52 &48& \textbf{100} &8& 40 &0& 36 &4& 20 &76& 72 & \textbf{96}& 76 \\
        Act3D & \textbf{27} & 51 & 82 & 90 & \textbf{75} & \textbf{95} & 26& 47 & \textbf{66} & 93 & 7& 8 & 6& 12 & 8& 9 &82& \textbf{92} & 64& \textbf{94} \\
        VIHE (ours) & 24 & \textbf{60} & \textbf{96} & \textbf{96} & 52 & 92 & \textbf{60} & \textbf{92} & 60 & 96 & \textbf{32} & \textbf{52} &\textbf{32}& \textbf{68} &\textbf{24}& \textbf{68} &\textbf{100}& 64 & 92 & 92 \\
        \bottomrule
    \end{tabular}
\end{table*}

\subsubsection{Action Prediction}
In initial stage 0, the transformer network $F$ outputs 8-dimensional action composing of $a_{pose}$, $a_{open}$ and $a_{col}$. In action prediction, we predict the 6-DoF $a_{pose}$ in SE(3) as a 3-DoF translation component $a_{trans}$ and a 3-DoF rotation component $a_{rot}$. As in PerAct~\cite{PerAct}, $a_{rot}$ is represented as Euler angles discretized into 5 degree bins. Following RVT~\cite{goyal2023rvt}, We first predict a heatmap for each view. Heatmaps from all views are then projected to predicted score for a set of 3D points to select $a_{trans}$ with the best score. Using the same set of heatmaps, we predict $a_{rot}, a_{open} \text{ and } a_{col}$ from image features weighted by the predicted translation heatmap. 

In subsequent refinement stages, the network outputs refinement transformation $h_{pose}$ in place of $a_{pose}.$ We note that, because the initial pose is from the base coordinate system, it is in fact a refinement from the world (identity) pose. We use the final stage outputs as the prediction of our model.

\subsubsection{Training}
Our model predicts later stage actions conditioned on the prediction from earlier stages. This makes our model inference autoregressive in nature. In autoregressive models, exposure bias~\cite{ranzato2015sequence} is known to deteriorate the performance due to the error accumulation from the discrepancy of predictions in training and inference. To mitigate this issue, we use stochastic sampling which adds a random perturbation when sampling camera poses for the refinement stages during training. Specifically, we add $\epsilon_i \in \text{SE}(3)$ to groundtruth $a_{pose}$ when rendering in-hand view images $x_i$ for stage $i$. This is different from testing where we use the predicted $a_{pose}^{i-1}$ to render in-hand view images as described in the previous subsection. 

Our final action prediction loss function is a summation of individual loss functions across all stages. Similar to PerAct~\cite{PerAct} and RVT~\cite{goyal2023rvt}, we choose to use cross-entropy loss for each component. For the continuous component $a_{trans}$, a target heatmap is obtained by applying a truncated Gaussian kernel to the 2D projection location of groundtruth 3D action as in RVT. The cross-entropy loss is then computed between the target heatmap and the predicted heatmap for each view. The final training loss $\mathcal{L}_{action}$ is then the summation of individual losses across all three stages as below.
\begin{align}
    \mathcal{L}_{action} &= \sum_{i=0}^2{
    [
        \mathcal{L}^i_{pose} + \mathcal{L}^i_{open} + \mathcal{L}^i_{col}
    ]}.
\end{align}

where $\mathcal{L}_{pose}^i = \mathcal{L}^i_{trans} + \mathcal{L}^i_{rot}$. ${L}^i_{trans}$, ${L}^i_{rot}$, ${L}^i_{open}$ and ${L}^i_{col}$ are losses for each component respectively.

\section{Experiments}

We evaluate \method in a multi-task, language-conditioned, vision-based imitation learning setting in both simulation and the real world. We conduct our simulated experiments in RLBench~\cite{RLBench2020}, an established simulation benchmark for learning manipulation policies. To validate its real-world applicability, we further test it on a physical Franka Panda arm~\cite{FrankaPanda}.

\subsection{Simulation Experiments} \label{sec:sim_exp}

\textbf{Simulation Setup.}
Our experimental framework closely aligns with the one established in PerAct~\cite{PerAct}. We utilize CoppeliaSim~\cite{VREP2013} to simulate a range of tasks from RLBench~\cite{RLBench2020}, employing a Franka Panda robot equipped with a parallel gripper. We assess performance across the same 18 tasks introduced by PerAct, which encompass a diverse set of activities such as pick-and-place, tool manipulation, and high-precision peg insertions. Each task is further diversified through variations guided by associated language descriptions. Visual observations are acquired from four \(128 \times 128\) resolution RGB-D cameras. To compute the target gripper pose, we employ a sampling-based motion planner~\cite{karaman2011sampling}, consistent with prior work~\cite{goyal2023rvt,gervet2023act3d,PerAct}.

\textbf{Baselines.}
We benchmark our approach against five established baselines: (1) Image-BC~\cite{BCZ}, which is an image-to-action behavior cloning agent with CNN and ViT vision encoder variants; (2) C2F-ARM-BC~\cite{C2FARM}, which transforms RGB-D images into multi-resolution voxels and predicts key-frame actions using a coarse-to-fine approach; (3) PerAct~\cite{PerAct}, which encodes RGB-D images into voxels and uses a Perceiver~\cite{PerceiverIO2021} transformer for action prediction; (4) RVT~\cite{goyal2023rvt}, which renders five global orthographic images from input RGB-D images and uses a multi-view transformer for action prediction; and (5) Act3D~\cite{gervet2023act3d}, which uses pre-trained image features and relative cross-attention on point clouds to detect actions.

\textbf{Implementation Details.}
We use the pre-trained ResNet-50\cite{he2016deep} variant of the CLIP model to acquire a sequence of \(K_{\text{lang}}=77\) tokens per instruction. We utilize the same image rendering pipeline introduced by RVT~\cite{goyal2023rvt}. RGB-D camera data is first projected into a point cloud, which is then used to render orthographic images from given viewpoints. Each rendered view consists of three image maps with a total of seven channels: 3 channel RGB, 1 channel depth, and 3 channels world-frame coordinates. All virtual images are rendered with a resolution of \(110 \times 110\). Each image is divided into \( K_{\text{img}} = 100\) patches of size \( 11 \times 11\). The model uses \( L = 8\) layers of masked self-attention, with an input of \( K_{\text{lang}} + 3\times 5\times K_{\text{img}} = 1577\) tokens. Notably, we render virtual views at half the resolution compared to RVT, yet achieving superior performance. This enhancement is attributable to the coarse-fine refinement strategy that capitalizes on the inherent multi-scale information in the images, facilitating more nuanced adjustments and optimization in the refinement stages.

\textbf{Training and Evaluation.}
We follow the training protocol that aligns with previous works~\cite{PerAct,goyal2023rvt} for fair benchmarking. We utilize the same RLBench dataset introduced by PerAct~\cite{PerAct}, along with the same data augmentation techniques. The model is trained for 150,000 steps using the LAMB optimizer~\cite{LAMB}, with a batch size of 24 and an initial learning rate of \( 2.4 \times 10^{-3} \). We train our method using the same hardware of 8 NVIDIA Tesla V100 GPUs as in previous works~\cite{PerAct, goyal2023rvt, gervet2023act3d}. The model is tested on the same 25 variations for each task as in PerAct~\cite{PerAct}. Benchmark evaluation in Table~\ref{tab:rlbench_main_ret} leverages results from respective sources for Image-BC~\cite{BCZ}, C2F-ARM-BC~\cite{C2FARM}, PerAct~\cite{PerAct}, and Act3D~\cite{gervet2023act3d}. We retrain RVT~\cite{goyal2023rvt} using author released code for fair comparison.

\begin{table}[!t]
    \caption{Ablation Study on Simulated Benchmark.}
    \label{table:ablation_study}
    \centering
    \setlength\tabcolsep{1.3pt}
    \begin{tabular}{lcc}
        \toprule
        Variants & Avg. Succ. & Perf. Change\\
        \midrule
        Full VIHE model & 77 & -\\
        
        \multicolumn{2}{l}{Network architectures} \\
        \quad No cross-stage attention & 69 & -8 \\
        \quad Direct predict $a_{pose}$ in refinement & 71 & -6 \\
        \quad No rotary positional encoding & 74 & -3 \\

        \multicolumn{2}{l}{Virtual in-hand view rendering} \\
        \quad No zoom-in for virtual in-hand views & 63 & -14 \\
        \quad Fixed rotation in virtual in-hand views & 69 & -8 \\
        \quad Position camera to look outside-in & 75 & -2 \\
        
        \multicolumn{2}{l}{Inference} \\
        \quad Use stage 0 action predictions for inference & 50 & -27 \\
        \quad Use stage 1 action predictions for inference & 71 & -6 \\        
        \bottomrule
    \end{tabular}
\end{table}

\textbf{Results.}
As presented in Table~\ref{tab:rlbench_main_ret}, \method surpasses all baselines in success rate when averaged across all tasks. With 100 demonstrations, it outperforms the existing SOTA method RVT by 17 percentage points (a 28\% relative improvement) and Act3D by 12 percentage points (an 18\% relative improvement). Our method's advantage sustains when only 10 demonstrations are provided, outperforming Act3D by 9 percentage points (a 19\% relative improvement). Remarkably, our performance using 10 demonstrations per task is close to the performance using 100 demonstrations per task for RVT. These results demonstrate that \method is both more accurate and more sample-efficient compared to existing state-of-the-art methods. The improvements mainly come from large gains in challenging high-precision tasks. For instance, \method greatly improves the success rate on the ``Insert Peg'' task from 27 to 84, more than tripling its performance and highlighting our method's capability in performing challenging high-precision tasks.

\textbf{Ablation Study.}
As shown in Table~\ref{table:ablation_study}, we conduct extensive ablation experiments on the same multi-task environments under 100 demonstrations. We analyze different design choices of \method, grouped into three categories: architecture, in-hand view rendering configurations, and model inference.

\textit{Architecture:}
We investigate the effect of different network components. Without cross-stage attention, the performance drops by 8\%. Directly predicting \( a_{\text{pose}} \) instead of predicting the relative transformation \( h_{\text{pose}} \) results in a performance decrease of 6\%. Rotary positional encoding also helps to efficiently relate information from patches in different stages, resulting in a 3\% difference.

\textit{Virtual In-Hand View Rendering:}
We vary camera settings when performing virtual in-hand view rendering. 
We found that iteratively decreasing the field of view by half in three stages is important, without which the performance drops to 63\%. Meanwhile, we also want to highlight that even without such iterative zoom-in, our model still largely outperforms single stage prediction, which achieves a 50\% success rate using images of our 100 resolution and 60\% using larger 220 resolution from RVT. This indicates that the in-hand pose itself, independent from iteratively zoom-in, offers strong inductive-bias for model to correlate image and action refinement.
Additionally, we also test rendering with fixed rotation and outside-in camera positions and find that they lead to performance drops of 8\% and 2\%, respectively. 

\textit{Inference:}
A model with only global observations at stage 0 suffers a 27\% performance drop, and a model with one additional refinement at stage 1 suffers a 6\% performance drop. This demonstrates that all three stages are important for achieving high performance.

\begin{figure}[h!]\label{fig:real_rollout}
  \centering 
  \includegraphics[width=3.4in]{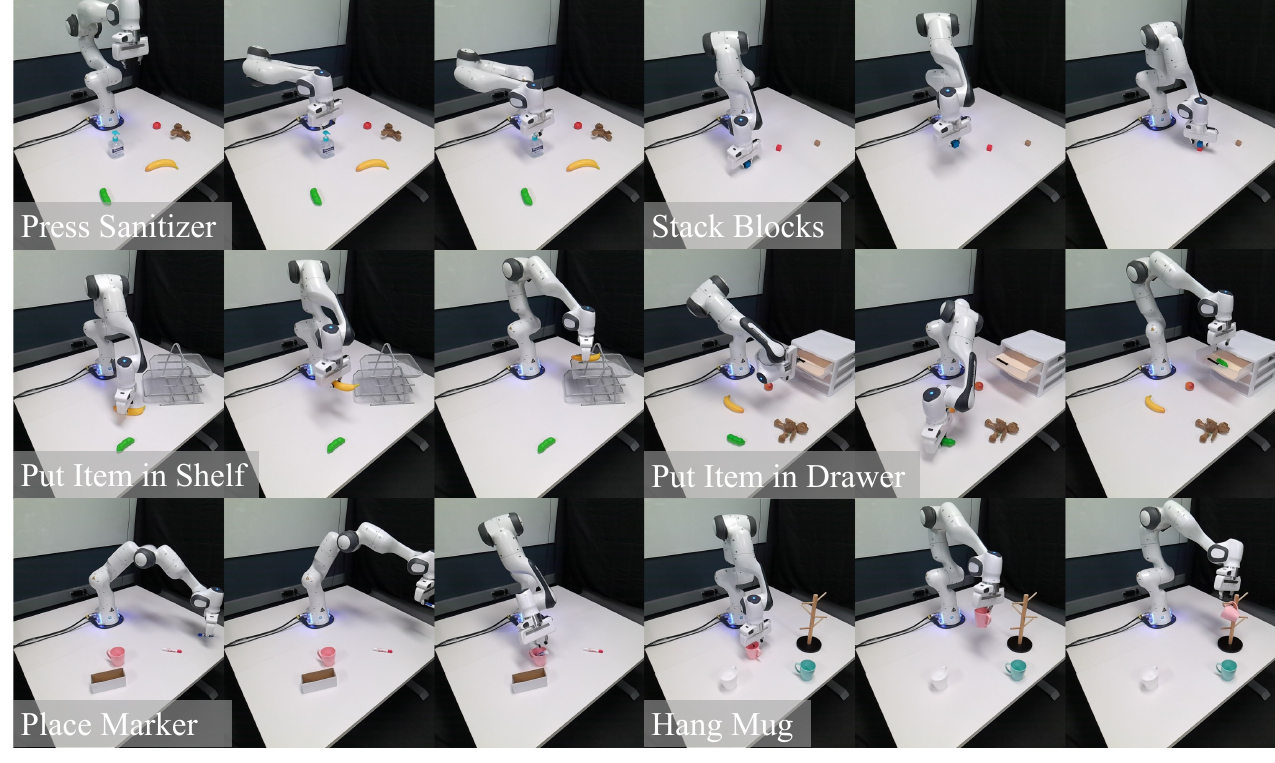}
  \caption{\textbf{Real-world object manipulation tasks.} A single \method{} model can perform multiple tasks (6 tasks, 18 variations) in the real world with just 72 demonstrations in total.}
\end{figure}

\subsection{Real-World Experiments}
\textbf{Real-World Setup.}
We experiment on a table-top setup using a Franka Panda arm~\cite{FrankaPanda}. The scene is perceived via an Azure Kinect (RGB-D) camera~\cite{AzureKinect} statically mounted in a third-person view. 
We design a total of 6 language-conditioned tasks with 18 variations, including different items, colors, and targets. A total of 72 demonstration trajectories are collected by executing human-specified waypoints with random locations of objects. 
Visual examples of our tasks can be found in Figure~\ref{fig:real_rollout}. More videos can be viewed on our project website.

\begin{table}[h]
  \caption{Multi-Task Configuration and Performance on Real Robot.}
  \label{table:real}
  \centering
  \setlength\tabcolsep{1.3pt}
\begin{tabular}{lcclcc}
\toprule
Task                & Variation           & \# Train~ & RVT Succ. & \method{} Succ.  \\
\midrule
{Press Sanitizer} & N/A (1)           &  5        & 9/10      & \textbf{10/10}        \\
{Stack Blocks}     & color (6)         & 12        & 4/10      & \textbf{7/10}        \\
{Put in Shelf}     & item  (2)         & 10        & 6/10      & \textbf{9/10}       \\
{Put in Drawer}    & item  (2)         & 10        & 3/10      & \textbf{5/10}       \\
{Place Marker}     & color and target (4) & 20   & 2/10      & \textbf{6/10}      \\
{Hang Mug}         & color (3)         & 15        & 1/10      & \textbf{7/10}        \\
\midrule
All tasks                 & 18                &  72       & 25/60      & \textbf{43/60}        \\
\bottomrule
\end{tabular}
\end{table}

\textbf{Results.} 
We train our \method and the baseline method RVT on real-world data for 50,000 steps, using the same hyper-parameters as in the simulated environment. Success rates across various tasks are provided in Table~\ref{table:real}. Overall, \method achieves a significant success rate of 43/60, outperforming RVT by a large margin. The results demonstrate our method's effectiveness in real-world 3D object manipulation tasks. Especially, the performance of \method stands out in tasks demanding high precision, such as ``Place Marker'' and ``Hang Mug''.  

\section{Conclusion}
In this work, we introduced a novel approach for robotic manipulation that leverages virtual in-hand views to iteratively refine action predictions. Our method offers an observation space that significantly enhances performance in 3D object manipulation. Through empirical evaluations, we showed that our approach substantially outperforms existing SOTA methods in both simulated environments and physical robots. In extensive ablation studies, we highlighted the importance of various design choices, including in-hand view rendering, relative transformation prediction, cross-stage attention, and zoom-in camera views.

We also identify some limitations that present avenues for future research. Similar to prior methods, VIHE requires calibrated RGB-D cameras to obtain point clouds for image rendering. Integrating our approach with NeRF-like implicit view rendering techniques~\cite{ze2023gnfactor, mildenhall2021nerf} may eliminate this constraint and broaden its applicability. Additionally, our current model does not utilize pre-trained image features, the inclusion of which may further improve performance.


\addtolength{\textheight}{-4.5cm}  
\bibliographystyle{IEEEtran}
\bibliography{IEEEabrv,bbl}
\addtolength{\textheight}{-10cm}  








\end{document}